\documentclass[sigconf]{acmart}

\usepackage{graphicx}
\usepackage{subcaption}
\usepackage{tabularx}
\usepackage{enumitem}
\usepackage{multirow}
\usepackage{algorithm}
\usepackage{algpseudocode}
\usepackage{fancyhdr}

\AtBeginDocument{%
  }

\renewcommand\footnotetextcopyrightpermission[1]{} 
\begin{document}

\title{PixelWeb: The First Web GUI Dataset with Pixel-Wise Labels}

\author{Qi Yang}
\affiliation{%
  \institution{Peking University}
  \city{Beijing}
  \country{China}}
\email{qi.yang@stu.pku.edu.cn}

\author{Weichen Bi}
\affiliation{%
  \institution{Peking University}
  \city{Beijing}
  \country{China}}
\email{biweichen@pku.edu.cn}

\author{Haiyang Shen}
\affiliation{%
  \institution{Peking University}
  \city{Beijing}
  \country{China}}
\email{hyshen@stu.pku.edu.cn}

\author{Yaoqi Guo}
\affiliation{%
  \institution{Peking University}
  \city{Beijing}
  \country{China}}
\email{ianwalls@pku.edu.cn}

\author{Yun Ma}
\affiliation{%
  \institution{Peking University}
  \city{Beijing}
  \country{China}}
\email{mayun@pku.edu.cn}

\renewcommand{\shortauthors}{Qi et al.}

\begin{abstract}

Graphical User Interface (GUI) datasets are crucial for various downstream tasks. However, GUI datasets often generate annotation information through automatic labeling, which commonly results in inaccurate GUI element BBox annotations, including missing, duplicate, or meaningless BBoxes. These issues can degrade the performance of models trained on these datasets, limiting their effectiveness in real-world applications. Additionally, existing GUI datasets only provide BBox annotations visually, which restricts the development of visually related GUI downstream tasks. To address these issues, we introduce PixelWeb, a large-scale GUI dataset containing over 100,000 annotated web pages. PixelWeb is constructed using a novel automatic annotation approach that integrates visual feature extraction and Document Object Model (DOM) structure analysis through two core modules: channel derivation and layer analysis. Channel derivation ensures accurate localization of GUI elements in cases of occlusion and overlapping elements by extracting BGRA four-channel bitmap annotations. Layer analysis uses the DOM to determine the visibility and stacking order of elements, providing precise BBox annotations. Additionally, PixelWeb includes comprehensive metadata such as element images, contours, and mask annotations. Manual verification by three independent annotators confirms the high quality and accuracy of PixelWeb annotations. Experimental results on GUI element detection tasks show that PixelWeb achieves performance on the mAP95 metric that is 3-7 times better than existing datasets. We believe that PixelWeb has great potential for performance improvement in downstream tasks such as GUI generation and automated user interaction.
\end{abstract}





\maketitle

\section{Introduction}

The increasing complexity and diversity of graphical user interfaces (GUIs) of web applications underscore the necessity for more precised GUI modeling, which is essential for various downstream tasks, such as GUI code generation~\cite{xiao2024prototype2code,wan2024automatically}, GUI retrieval~\cite{bunian2021vins}, and user interaction automation through LLM-based agents~\cite{wan2024omniparser,you2024ferret,li2024ferret}. 

\begin{figure}[htbp]
    \centering
    \begin{subfigure}[t]{0.15\textwidth}
        \centering
        \includegraphics[width=\textwidth]{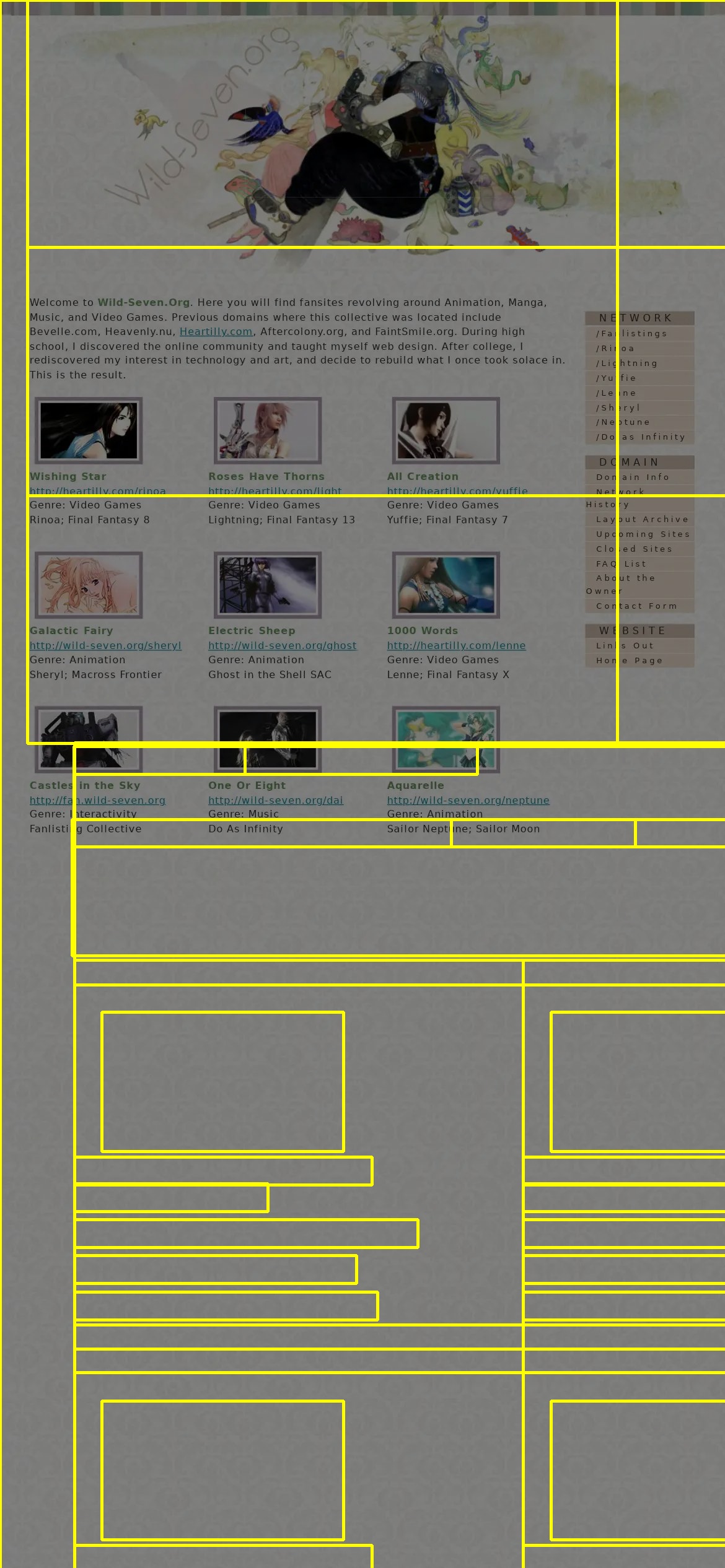}
        \caption{The positions of the BBoxes do not have GUI elements}
        \label{fig:error_webui_1}
    \end{subfigure}
    \hfill
    \begin{subfigure}[t]{0.15\textwidth}
        \centering
        \includegraphics[width=\textwidth]{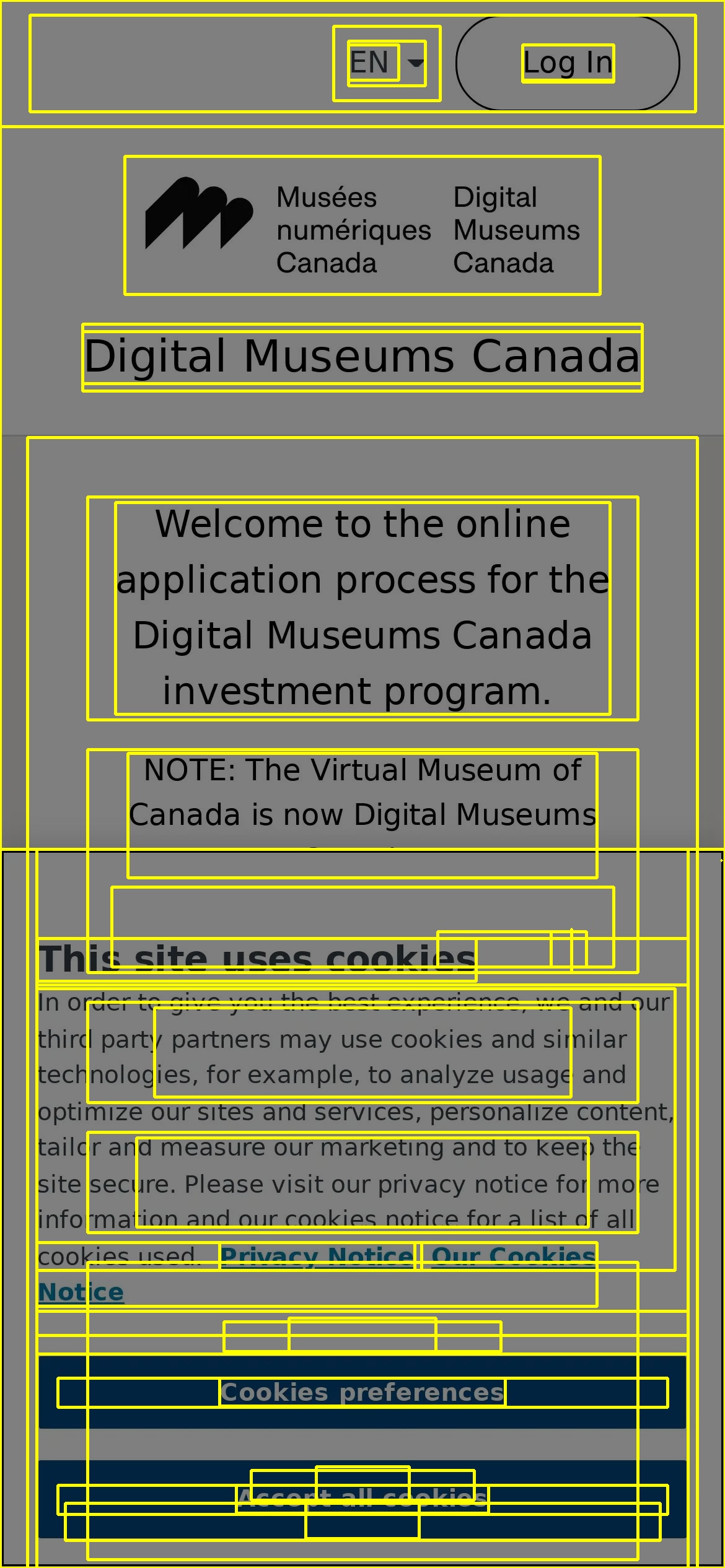}
        \caption{The same element corresponds to multiple BBoxes.}
        \label{fig:error_webui_3}
    \end{subfigure}
    \hfill
    \begin{subfigure}[t]{0.15\textwidth}
        \centering
        \includegraphics[width=\textwidth]{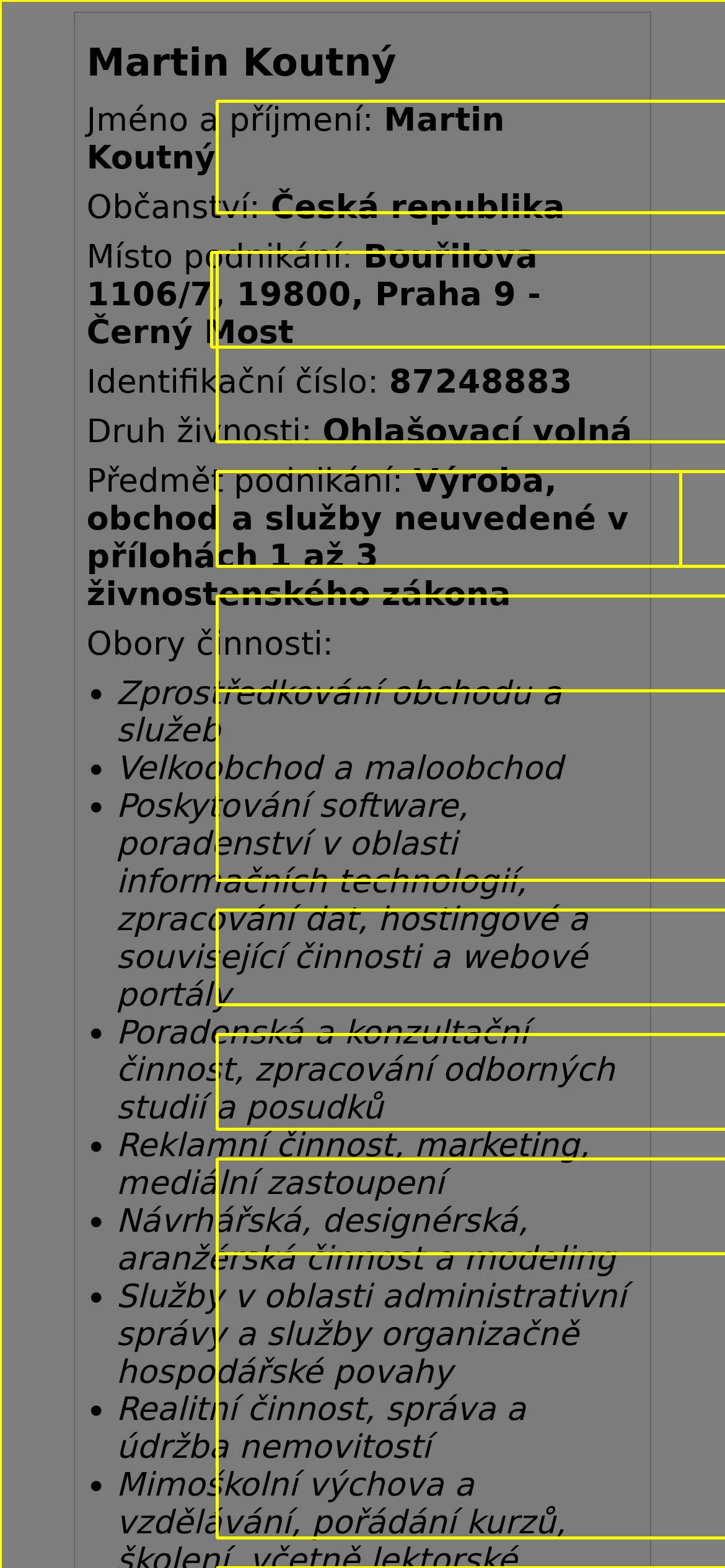}
        \caption{The positions of the BBoxes are offset relative to the GUI elements.}
        \label{fig:error_webui_2}
    \end{subfigure}

    \caption{Error cases of BBox label in WebUI dataset}
    \label{fig:error_demo_comparison}
\end{figure}

Precise GUI modeling relies on high-quality and large-scale datasets for model training. However, existing widely-used GUI datasets, such as WebUI~\cite{Wu2023WebUI}, Rico~\cite{deka2017rico}, and MUD~\cite{Feng2024MUD}, suffer from pervasive annotation inaccuracies~\cite{li2022learning}. Figure~\ref{fig:error_demo_comparison} shows some cases of imprecise BBox labels in WebUI dataset: (a) some BBoxes do not have corresponding GUI elements; (b) the same GUI element corresponds to multiple BBoxes; (c) the positions of some BBoxes do not align with the GUI elements. These annotation issues can introduce noise and undermine the reliability of trained models, adversely affecting the performance of downstream tasks.

Addressing these limitations faces two primary challenges: (1) Unknown Accurate Coordinates of Elements: Existing browser APIs for obtaining element coordinates often fail to accurately reflect the actual positions and sizes of GUI elements~\cite{li2022learning}, especially for non-rectangular and dynamically rendered components. This leads to incorrect BBox sizes and imprecise localization; (2) Unknown Visibility of Elements: Determining the visibility of elements is complicated by factors such as overlapping components and varying display conditions~\cite{li2022learning}. Simple code analysis cannot reliably determine whether elements are visible to users, resulting in missing, duplicate, or meaningless BBoxes.

To this end, we propose an automated annotation approach designed to enhance the accuracy of GUI dataset construction. Our approach leverages the synergy between visual features and DOM structure through two core modules: \textit{Channel Derivation} and \textit{Layer Analysis}. Channel Derivation extracts the BGRA channels for each pixel of GUI elements, ensuring accurate localization even in complex scenarios involving occlusions and overlapping elements. Layer Analysis examines the element hierarchy by analyzing the DOM tree and z-index information, determining the visibility and stacking order of elements. Together, these modules generate precise BBox annotations and comprehensive metadata with minimal human intervention.

Based on this approach, we construct \textit{\textbf{PixelWeb}}~\footnote{PixelWeb can be accessed from https://huggingface.co/datasets/cyberalchemist/PixelWeb}, a Web GUI dataset comprising \textbf{100,000} annotated web pages. PixelWeb offers more precise BBox annotations and more metadata labels of each element such as image, mask, and contour. Experimental results demonstrate that models trained on PixelWeb significantly outperform those trained on existing datasets, highlighting the efficacy of our approach in producing high-quality annotations.

Experimental evaluations demonstrate the efficacy of approach in producing high-quality annotations. Manual verification conducted by three independent annotators revealed higher annotation quality for BBox annotations and more comprehensive metadata. Furthermore, experimental results indicate that element detection models built using PixelWeb significantly outperform existing GUI datasets in detection accuracy. This implies that PixelWeb can play a greater role in more downstream tasks based on element detection, such as GUI operations and understanding.

In summary, our contributions are as follows.
\begin{itemize}
\item \textbf{Pixel-wise Annotation Approach for Web GUI}: We propose an automated annotation approach that integrate visual features and DOM structure to overcome existing annotation challenges.

\item \textbf{PixelWeb Dataset}: We construct a large-scale, automated GUI dataset encompassing 100,000 web pages with more precise BBox annotations, as well as more metadata labels, such as the image, mask and contour of each element.

\item \textbf{Enhanced Performance of Downstream Tasks}: We validate the effectiveness of PixelWeb through extensive experiments, demonstrating significant improvements in GUI element detection, thereby underscoring the dataset's value for more downstream tasks.
\end{itemize}

\section{Related Work}
In this section, we survey related work on GUI datasets and downstream tasks enabled by GUI datasets.

\subsection{GUI Datasets}
\label{sec:GUI_Datasets}
The construction of GUI datasets has been extensively explored to support downstream tasks in UI analysis and automation. Some research improves downstream task performance by constructing datasets containing diverse GUI-related information. For instance, Rico~\cite{deka2017rico} is a seminal mobile UI repository constructed by combining crowdsourcing and automation to extract visual, structural, and interactive properties from thousands of apps. VINS~\cite{bunian2021vins} is a dataset with hierarchical UI structure annotations to enable object-detection-based retrieval. UICrit~\cite{duan2024uicrit} focuses on enhancing automated UI evaluation by curating a critique dataset to refine LLM-generated feedback, bridging the gap between automated and human evaluators. WebUI~\cite{Wu2023WebUI} crawls web pages to create a large-scale dataset with web semantics for cross-domain visual UI understanding. Some work focuses on automating dataset acquisition to reduce manual hard labor. For example, CLAY~\cite{li2022learning} proposes a deep learning pipeline to denoise raw mobile UI layouts, automating dataset refinement and reducing manual labeling efforts. MUD~\cite{Feng2024MUD} employs LLMs to mine modern UIs from apps, integrating noise filtering and human validation to address outdated or noisy data. However, previous work faced inherent limitations in annotation precision (e.g., noisy BBoxes, overlapping elements) or scalability due to manual interventions. Our approach proposes pixel-level precise annotations, which helps further enhance the performance of downstream tasks.

\subsection{GUI Tasks}
\label{sec:GUI_Tasks}

Numerous studies have focused on advancing performance in GUI-related tasks. Some work focuses on GUI element detection: UIED~\cite{xie2020uied} combines traditional CV and deep learning for element detection, and Chen et al.~\cite{chen2020object} merge coarse-to-fine strategies with deep learning. Another task is GUI retrieval. For instance, Guigle~\cite{bernal2019guigle} helps conceptualize the user interfaces of the app by indexing GUI images and metadata for intuitive search. In layout generation, approachs like BLT~\cite{kong2022blt} and LayoutDM~\cite{inoue2023layoutdm} improve controllability and quality, while LayoutTransformer~\cite{gupta2021layouttransformer} unifies cross-domain layout synthesis. Furthermore, GUI agents such as AutoGLM~\cite{liu2024autoglm} integrate reinforcement learning for autonomous control. Compared to previous work, our approach leverages a pixel-level annotated dataset (PixelWeb), enabling enhanced performance across GUI tasks.

\section{Approach}
\begin{figure*}[t]
    \centering
    \includegraphics[width=\textwidth]{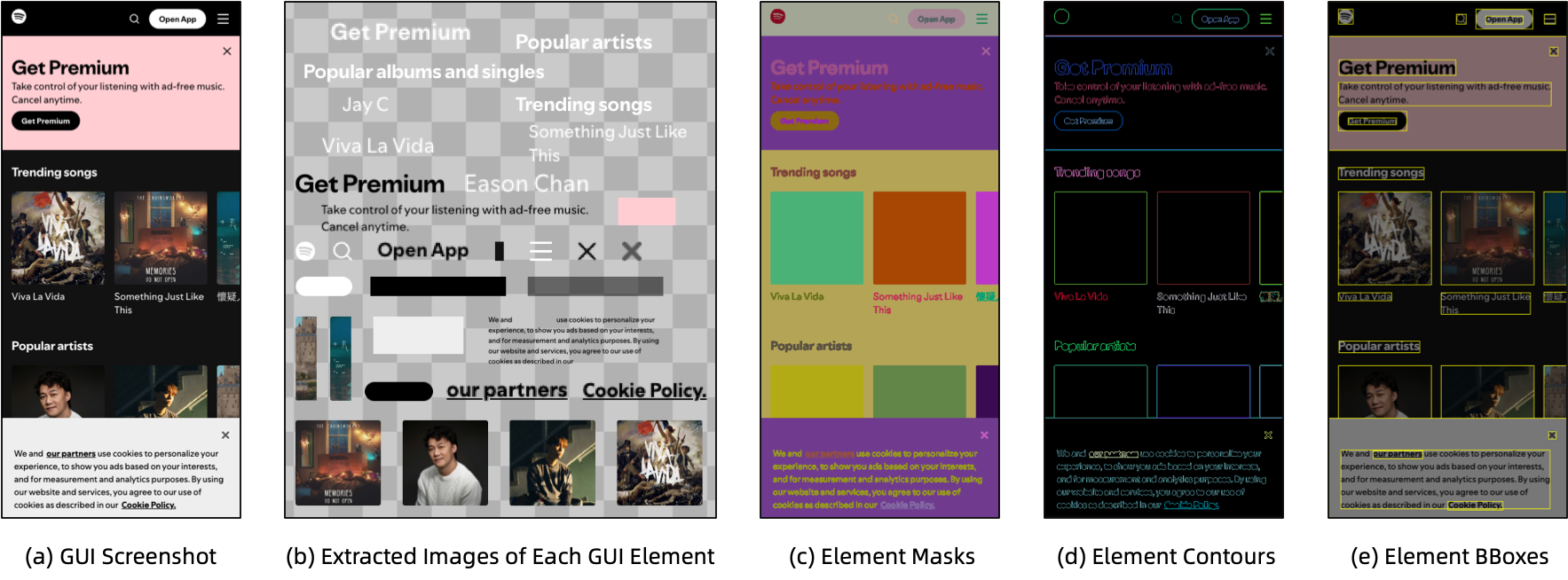}
    \caption{Example of a web page annotated by our approach}
    \label{fig:spotify_demo}
\end{figure*}

In this section, we present an automated annotation approach for Web GUIs that enables large-scale dataset construction, which delivers (1) relatively precise BBox annotations and (2) more metadata labels, including the image, mask, contour of each GUI element. The approach consists of two core modules: (1) \textit{Channel Derivation} (Section \ref{subsec:4.2}), which precisely extracts the image of each GUI element, and (2) \textit{Layer Analysis} (Section \ref{subsec:4.3}), which examines element hierarchy as well as produces lower-dimensional annotation (masks, contours, and BBoxes).

\begin{table}
\caption{Annotation differences among datasets}
\label{tab:datasets_diff}
\small
\begin{tabularx}{0.48\textwidth}{l|cccc}

\hline
Features & PixelWeb & WebUI~\cite{Wu2023WebUI} & Rico~\cite{deka2017rico}
& MUD~\cite{Feng2024MUD} \\ \hline
Page Screenshot & \checkmark & \checkmark & \checkmark & \checkmark \\
Page Hierarchy Code & \checkmark & \checkmark & \checkmark & \checkmark \\
Page Animation &  &  & \checkmark &  \\
Element Image & \checkmark &  &  &  \\
Element Layer & \checkmark &  &  &  \\
Element Mask & \checkmark &  &  &  \\
Element Contour & \checkmark &  &  &  \\
Element BBox & \checkmark & \checkmark & \checkmark & \checkmark \\
Element Class & \checkmark & \checkmark &  & \checkmark \\
Element Computed Style & \checkmark &  &  &  \\
\hline
\end{tabularx}
\end{table}

\subsection{Overview}
\label{subsec:4.1}

Achieving precise automated annotation of GUI elements from web pages faces two technical challenges:

\begin{itemize}
    \item \textbf{Unknown Accurate Coordinates of Elements}: Although browsers provide various APIs to obtain element coordinates, they cannot accurately reflect the actual positions of elements, leading to incorrect BBox sizes.
    
    \item \textbf{Unknown Visibility of Elements}: The visibility of elements is influenced by multiple factors, making it difficult to determine whether elements are visible to users through simple code analysis. This can result in missing BBoxes, duplicate BBoxes, and meaningless BBoxes.
\end{itemize}

\begin{figure*}[t]
    \centering
    \includegraphics[width=\textwidth]{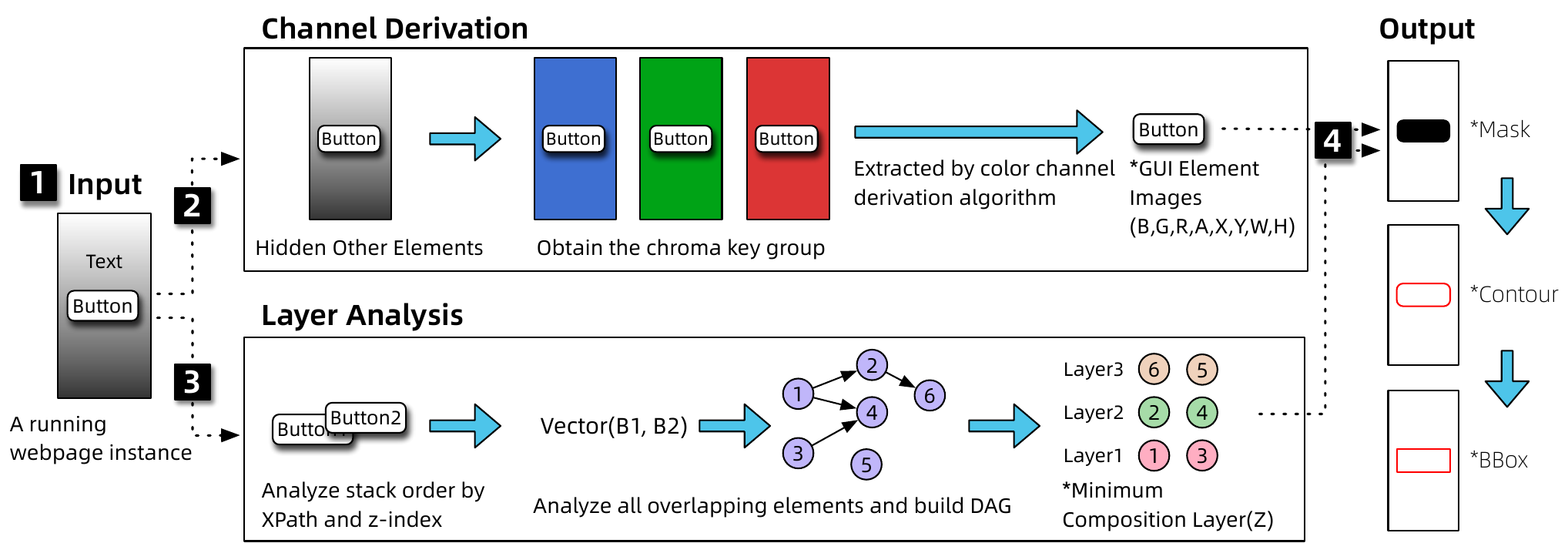}
    \caption{Approach overview. 1. Input an open web page. 2. The channel derivation module extracts the image of each element. 3. The layer analysis analyzes the rendering layer of elements to determine the image and coordinates of each GUI element. 4. Based on this information, it sequentially derives mask, contour, and BBox annotations.}
    \label{fig:approach_overview}
\end{figure*}

To address these challenges, we designed two modules: channel derivation and layer analysis. An overview of our approach is shown in Figure \ref{fig:approach_overview}, which consists of three steps:

\begin{enumerate}
    \item Open the target webpage to be annotated.
    \item Obtain the BGRA four-channel bitmap and XY coordinates of each GUI element through \textbf{Channel Derivation}.
    \item Determine the layer position (i.e., the Z-axis coordinate) of each GUI element through \textbf{Layer Analysis}.
\end{enumerate}

Using the information obtained from these modules, we can further derive image annotation data such as masks, contours, and BBoxes.

\textbf{Channel Derivation}: By independently displaying the web elements to be annotated and changing the page's background color, a chroma key group can be obtained. This allows us to solve the color composition equations to obtain the BGRA four-channel bitmap of each web element.

\textbf{Layer Analysis}: Based on the webpage's XPath and z-index information, we construct vectors that describe the stacking order of overlapping elements on the page. This results in several directed acyclic graphs (DAGs), which are then analyzed using a topological sorting algorithm to determine the hierarchy of elements.

After processing with these two modules, we obtain the image and spatial coordinates of each element. Using this high-dimensional information, we can derive lower-dimensional annotation data. Table \ref{tab:datasets_diff} illustrates the differences in annotations provided by our dataset compared to others. Our dataset offers more comprehensive information at the element level. Although we do not include the Animation annotations available in the Rico dataset, the Rico dataset requires manual annotation, and our targeted WebUI dataset also does not provide such annotations.

Figure \ref{fig:spotify_demo} displays a webpage annotated using our approach. The primary distinction between our GUI dataset and previous ones is the additional \textit{element images} label shown in Figure \ref{fig:spotify_demo}b. By combining this with the rendering hierarchy, we can compute annotation information such as the mask(\ref{fig:spotify_demo}c), contour\ref{fig:spotify_demo}d, and BBox(\ref{fig:spotify_demo}e) of elements.

\subsection{Channel Derivation}
\label{subsec:4.2}
\begin{figure}[]
    \centering
    \begin{subfigure}[b]{\columnwidth}
        \centering
        \includegraphics[width=\textwidth]{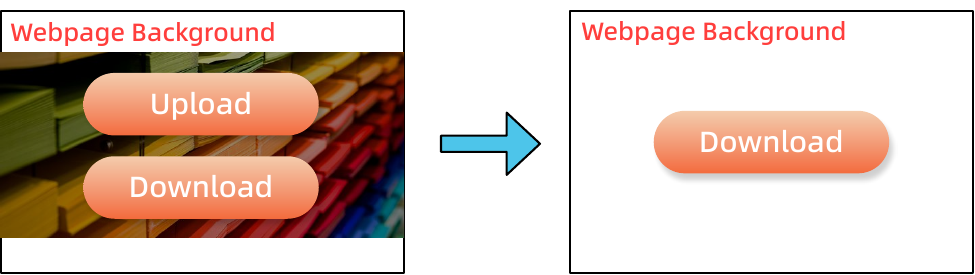}
        \caption{Exclude interfering elements. Hide all elements except the target elements to be extracted}
        \label{fig:approach_stage1_a}
    \end{subfigure}

    \vspace{2em} 
            
    \begin{subfigure}[b]{\columnwidth}
        \centering
        \includegraphics[width=\textwidth]{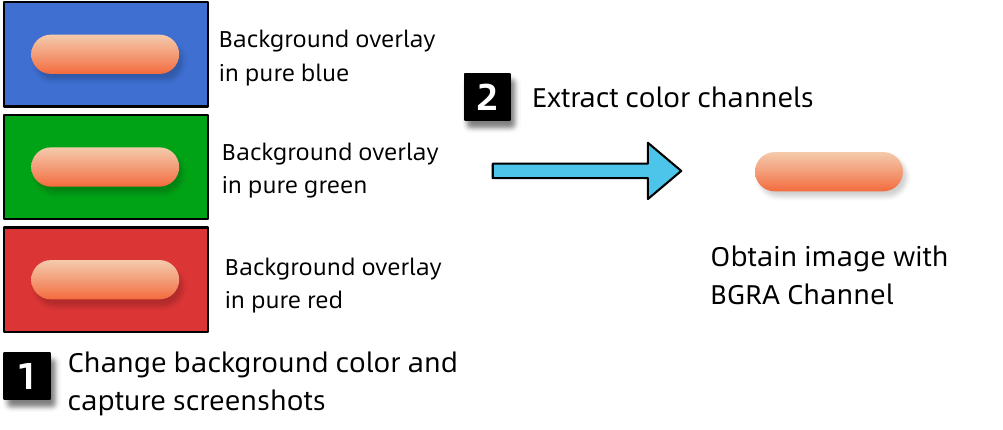}
        \caption{Extract graphic color channel: The three GUI screenshots on the left from top to bottom are: background set to blue, background set to green, background set to red}
        \label{fig:approach_stage1_b}
    \end{subfigure}

    \vspace{2em} 
        
    \begin{subfigure}[b]{\columnwidth}
        \centering
        \includegraphics[width=\textwidth]{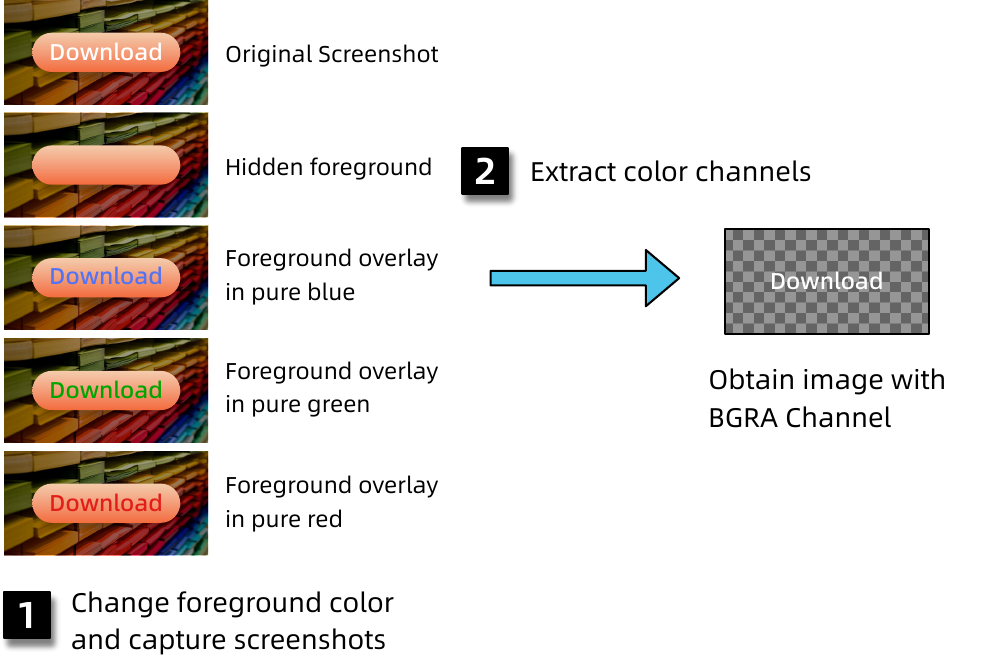}
        \caption{Extract text color channel: The five GUI screenshots on the left from top to bottom are: no changes, text removed, text set to blue, text set to green, text set to red. Because the font is white, we overlay a checkerboard pattern background for better visualization.}
        \label{fig:approach_stage1_c}
    \end{subfigure}
    
    \caption{Channel derivation}
    \label{fig:approach_stage1}
\end{figure}

Extracting GUI element images directly from GUI screenshots is a challenging task due to various special cases that make the direct use of simple cropping and contour extraction algorithms infeasible. For instance, elements may be occluded, leading to some regions being invisible. Additionally, the boundaries between elements may be blurred, making it difficult to accurately delineate areas. Furthermore, some elements may have the same color as the background, making it challenging to distinguish between the elements themselves and the page background.

Therefore, we designed a approach to extract GUI element images as follows: First, hide all elements except the target element, then change the background color of the HTML page to obtain screenshots with different chroma keys. This allows us to derive the BGRA color channel values of any pixel of the target element. We specifically designed a color derivation algorithm for this purpose. Using the standard color composition formula\ref{eq:color_render}:

\begin{equation}
\label{eq:color_render}
\text{V}_{\text{rendered}} = \alpha \times \text{V}_{\text{foreground}} + (1-\alpha) \times \text{V}_{\text{background}}
\end{equation}

we can obtain GUI screenshots under different background overlays. Through these screenshots and the color composition formula, we deduce the BGRA channel values of each pixel of the foreground GUI element.

Due to the characteristics of HTML syntax, a hierarchical tree node may contain more than one visual object. Although we can hide or display any HTML node, it is not possible to simply set a node’s text to be visible while keeping the background hidden. This prevents us from obtaining a text chroma key group by overlaying backgrounds of different colors. Therefore, we designed two extraction approach, applicable to situations where the background can be removed and where it cannot. For cases where the background can be removed, we perform a color transformation on the background to obtain the chroma key group. In situations where the background cannot be removed, we perform a color transformation on the foreground to obtain the chroma key group. In the following text, we refer to the former image as a graphic and the latter as text.

We designed different processing workflows for the above two situations, as shown in Figure \ref{fig:approach_stage1}. Regardless of the situation, we first hide all other irrelevant elements, retaining only the background of the HTML itself \ref{fig:approach_stage1_a}. For graphic extraction \ref{fig:approach_stage1_b}, we independently display it and change the background color to (255,0,0), (0,255,0), (0,0,255) to obtain the chroma key group. We solve for the BGRA channel values using the corresponding color channel derivation formula. For text extraction \ref{fig:approach_stage1_c}, we capture a screenshot retaining the element and one without the element to obtain the element's mask. We then change its own color and use channel derivation to obtain its BGRA channels.

\begin{table}[t] 
\caption{List of variables and descriptions}
\label{tab:variables}
\centering
\begin{tabularx}{0.47\textwidth}{|c|X|}
\hline
    
\textbf{Variable} & \textbf{Description} \\ \hline

\(\epsilon\) & A very small value to prevent division by zero \\ \hline

\((B,G,R,A)_{\text{FG}}\) & Graphic color channels \\ \hline
\((B,G,R,A)_{\text{FG}}'\) & Text color channels \\ \hline
\((B,G,R)_{\text{BG}}\) & Graphic background color channels \\ \hline
\((B,G,R)_{\text{BG}}'\) & Text background color channels \\ \hline

\((B,G,R)_{\text{BlueBG}}\) & Color channel values of GUI screenshot with blue background overlay \\ \hline
\((B,G,R)_{\text{GreenBG}}\) & Color channel values of GUI screenshot with green background overlay \\ \hline
\((B,G,R)_{\text{RedBG}}\) & Color channel values of GUI screenshot with red background overlay \\ \hline

\(A_{\text{(Blue,Green,Red)}}'\) & Alpha channel values calculated from different color channels \\ \hline

\(Mask_{\text{WhiteBG}}\) & Area where the background is pure white \\ \hline

\((B,G,R)_{\text{BlueFG}}\) & Color channel values of a GUI screenshot when the text is set to blue \\ \hline
\((B,G,R)_{\text{GreenFG}}\) & Color channel values of a GUI screenshot when the text is set to green \\ \hline
\((B,G,R)_{\text{RedFG}}\) & Color channel values of a GUI screenshot when the text is set to red \\ \hline
\((B,G,R)_{\text{RawFG}}\) & Color channel values of a GUI screenshot when the text is keep unchanged \\ \hline

\end{tabularx}
\end{table}

\begin{figure}[]
    \centering

    \begin{subfigure}[b]{\columnwidth}
        \centering
        \includegraphics[width=\textwidth]{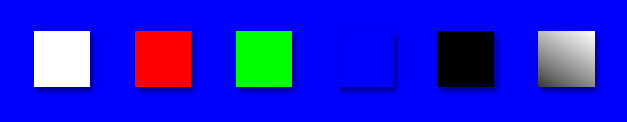}
        \caption{Input 1: Blue background}
        \label{fig:bgra_sample_graphic_blue}
    \end{subfigure}

    \vspace{0.5em} 
  
    \begin{subfigure}[b]{\columnwidth}
        \centering
        \includegraphics[width=\textwidth]{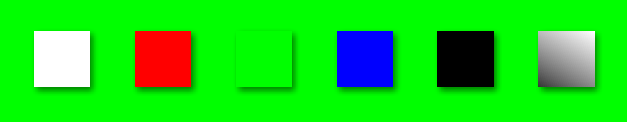}
        \caption{Input 2: Red background}
        \label{fig:bgra_sample_graphic_green}
    \end{subfigure}

    \vspace{0.5em} 

        \begin{subfigure}[b]{\columnwidth}
        \centering
        \includegraphics[width=\textwidth]{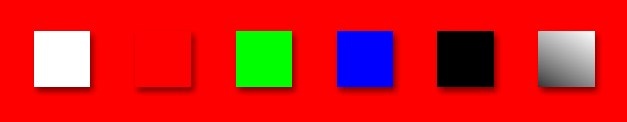}
        \caption{Input 3: Red background}
        \label{fig:bgra_sample_graphic_red}
    \end{subfigure}

    \vspace{0.5em} 

    \begin{subfigure}[b]{\columnwidth}
        \centering
        \includegraphics[width=\textwidth]{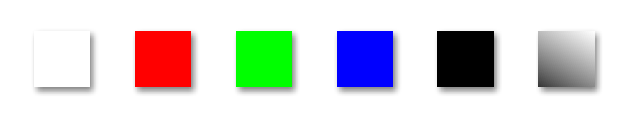}
        \caption{Output: BGRA channels of foreground}
        \label{fig:bgra_sample_graphic_derived}
    \end{subfigure}

    \caption{Examples of extreme cases for graphic color channel derivation formula}
    \label{fig:bgra_example_graphic}
\end{figure}

\begin{figure}[]
    \centering

    \begin{subfigure}[b]{\columnwidth}
        \centering
        \includegraphics[width=\textwidth]{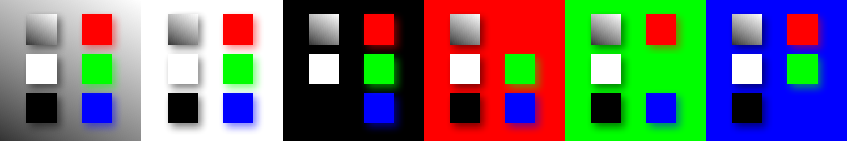}
        \caption{Input 1: Original GUI Screenshot}
        \label{fig:bgra_sample_text_orgin}
    \end{subfigure}

    \vspace{0.5em} 

    \begin{subfigure}[b]{\columnwidth}
        \centering
        \includegraphics[width=\textwidth]{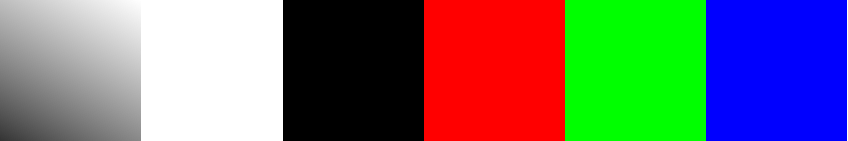}
        \caption{Input 2: Hidden foreground}
        \label{fig:bgra_sample_text_bg}
    \end{subfigure}

    \vspace{0.5em} 

    \begin{subfigure}[b]{\columnwidth}
        \centering
        \includegraphics[width=\textwidth]{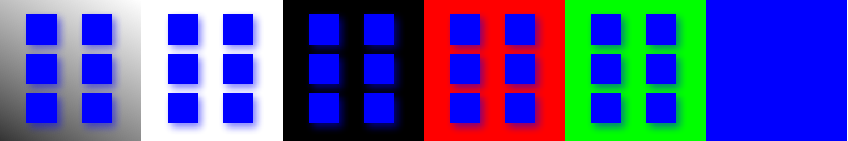}
        \caption{Input 3: Blue foreground}
        \label{fig:bgra_sample_text_blue}
    \end{subfigure}

    \vspace{0.5em} 

        \begin{subfigure}[b]{\columnwidth}
        \centering
        \includegraphics[width=\textwidth]{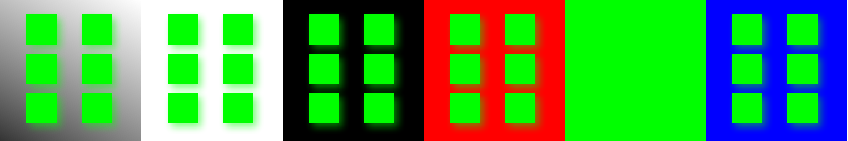}
        \caption{Input 4: Green foreground}
        \label{fig:bgra_sample_text_green}
    \end{subfigure}

    \vspace{0.5em} 

        \begin{subfigure}[b]{\columnwidth}
        \centering
        \includegraphics[width=\textwidth]{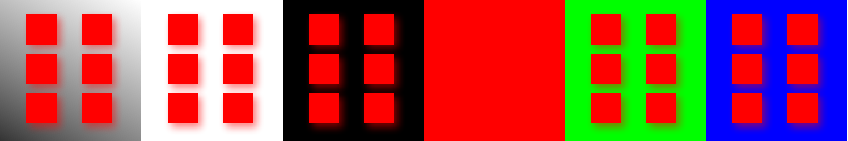}
        \caption{Input 5: Red foreground}
        \label{fig:bgra_sample_text_red}
    \end{subfigure}

    \vspace{0.5em} 
            
    \begin{subfigure}[b]{\columnwidth}
        \centering
        \includegraphics[width=\textwidth]{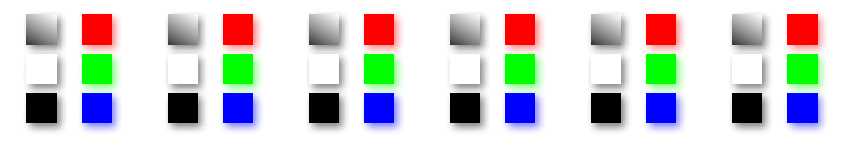}
        \caption{Output: BGRA channels of foreground}
        \label{fig:bgra_sample_text_derived}
    \end{subfigure}

    \caption{Examples of extreme cases for text color channel derivation formula}
    \label{fig:bgra_example_text}
    
\end{figure}

In the detailed description of the formulas that follow, we refer to the object to be extracted from the GUI screenshot as the foreground, and its background as the background. On this point, we consider all content other than the target to be extracted as the background, even if there are multiple layers, treating them as a single background layer. Whether extracting graphics or text, in their respective chroma key groups, they belong to the foreground, and anything other than themselves is considered the background. For the range of color channels, we use the standard [0,255] for the B, G, R channels, and [0,1] for the alpha channel.

Figures~\ref{fig:bgra_example_graphic} and \ref{fig:bgra_example_text} are examples of extreme cases of the bgra channel derivation formula and show the inputs and outputs of the formula. We use different extreme colors to test the formula, including pure white, pure black, gradient gray, pure red, pure green, and pure blue. Pure white and pure black represent the maximum and minimum values of all color channels, respectively; pure red, pure green, and pure blue represent the maximum values of each color channel in the RGB color space; gradient gray is used to test the formula's ability to handle mid-tone pixels.

Figure~\ref{fig:bgra_example_graphic} shows examples of graphic extraction. Figure~\ref{fig:bgra_sample_graphic_blue} shows a screenshot when the GUI background is set to pure blue, Figure~\ref{fig:bgra_sample_graphic_green} shows a screenshot when the GUI background is set to pure green, Figure~\ref{fig:bgra_sample_graphic_red} shows a screenshot when the GUI background is set to pure red, and Figure~\ref{fig:bgra_sample_graphic_derived} shows the output obtained through the graphic bgra channel derivation formula from Figures~\ref{fig:bgra_sample_graphic_blue}, \ref{fig:bgra_sample_graphic_green}, \ref{fig:bgra_sample_graphic_red}.

Figure~\ref{fig:bgra_example_text} shows examples of text extraction. Figure~\ref{fig:bgra_sample_text_orgin} shows the original GUI screenshot, Figure~\ref{fig:bgra_sample_text_bg} shows the GUI screenshot after hiding the text, Figure~\ref{fig:bgra_sample_text_blue} shows a screenshot when the text is set to pure blue, Figure~\ref{fig:bgra_sample_text_green} shows a screenshot when the text is set to pure green, Figure~\ref{fig:bgra_sample_text_red} shows a screenshot when the text is set to pure red, and Figure~\ref{fig:bgra_sample_text_derived} shows the output obtained through the text bgra channel derivation formula from Figures~\ref{fig:bgra_sample_text_orgin}, \ref{fig:bgra_sample_text_bg}, \ref{fig:bgra_sample_text_blue}, \ref{fig:bgra_sample_text_green}, \ref{fig:bgra_sample_text_red}.

First, we will introduce the extraction of graphics. Using the standard formula for color composition \ref{eq:color_render}, we can list the system of equations shown in formula \ref{eq:graphic_base_equation_group}.

\begin{equation}
\label{eq:graphic_base_equation_group}
\left\{
\begin{aligned}
(B, G, R)_{\text{BlueBG}} &= A_{\text{FG}} \cdot (B, G, R)_{\text{FG}} + (1 -  A_{\text{FG}}) \cdot (255, 0, 0) \\
(B, G, R)_{\text{GreenBG}} &= A_{\text{FG}} \cdot (B, G, R)_{\text{FG}} + (1 -  A_{\text{FG}}) \cdot (0, 255, 0) \\
(B, G, R)_{\text{RedBG}} &= A_{\text{FG}} \cdot (B, G, R)_{\text{FG}} + (1 -  A_{\text{FG}}) \cdot (0, 0, 255)
\end{aligned}
\right.
\end{equation}

We can first solve for the value of the A channel of the graphic using the system of equations \ref{eq:graphic_base_equation_group}.

\begin{align}
\label{eq:alpha}
A_{\text{FG}} = 1 - \frac{1}{6} \Bigg( & \frac{|B_\text{BlueBG} - B_\text{RedBG}|}{255} + \frac{|B_\text{BlueBG} - B_\text{GreenBG}|}{255} \nonumber \\
& + \frac{|G_\text{GreenBG} - G_\text{BlueBG}|}{255} + \frac{|G_\text{GreenBG} - G_\text{RedBG}|}{255} \nonumber \\
& + \frac{|R_\text{RedBG} - R_\text{GreenBG}|}{255} + \frac{|R_\text{RedBG} - R_\text{BlueBG}|}{255} \Bigg)
\end{align}

Finally, we can substitute the value of the A channel into formula \ref{eq:color_render} to obtain the expressions for the blue, green, and red channel values \ref{eq:graphic_bgr_equation_group}.

\begin{equation}
\label{eq:graphic_bgr_equation_group}
\left\{
\begin{aligned}
B_{\text{FG}} &= \frac{B_{\text{BlueBG}} - (1 - A_{\text{FG}}) \cdot 255}{A_{\text{FG}} + \epsilon} \\
G_{\text{FG}} &= \frac{G_{\text{GreenBG}} - (1 - A_{\text{FG}}) \cdot 255}{A_{\text{FG}} + \epsilon} \\
R_{\text{FG}} &= \frac{R_{\text{RedBG}} - (1 - A_{\text{FG}}) \cdot 255}{A_{\text{FG}} + \epsilon}
\end{aligned}
\right.
\end{equation}

Next, we will detail the extraction of text. For the set of GUI screenshots obtained by setting different color values, solving for the BGRA of text elements becomes more challenging. This is because the background of text nodes can be any background, making both the background channel values and the foreground (text) channel values unknown. We can first list the equations as follows:

\begin{equation}
\label{eq:text_base_equation_group}
\left\{
\begin{aligned}
(B, G, R)_{\text{BlueFG}} &= A_{\text{FG}}' \cdot (B, G, R) + (1 - A_{\text{FG}}') \cdot (B, G, R)_{\text{BG}}' \\
(B, G, R)_{\text{GreenFG}} &= A_{\text{FG}}' \cdot (B, G, R) + (1 - A_{\text{FG}}') \cdot (B, G, R)_{\text{BG}}' \\
(B, G, R)_{\text{RedFG}} &= A_{\text{FG}}' \cdot (B, G, R) + (1 - A_{\text{FG}}') \cdot (B, G, R)_{\text{BG}}'
\end{aligned}
\right.
\end{equation}

Based on the system of equations \ref{eq:text_base_equation_group}, we can first solve the system of equations \ref{eq:text_alpha_equation_group} that describes the foreground alpha channel:

\begin{equation}
\label{eq:text_alpha_equation_group}
\left\{
\begin{aligned}
A_{\text{Blue}}' &= \frac{B_{\text{BlueFG}} - B_{\text{BG}}'}{255 - B_{\text{BG}}' + \epsilon} \\
A_{\text{Green}}' &= \frac{G_{\text{GreenFG}} - G_{\text{BG}}'}{255 - G_{\text{BG}}' + \epsilon} \\
A_{\text{Red}}' &= \frac{R_{\text{RedFG}} - R_{\text{BG}}'}{255 - R_{\text{BG}}' + \epsilon}
\end{aligned}
\right.
\end{equation}

Since we calculate alpha from three different color channels, and there is only one final alpha channel, we use equation \ref{eq:text_alpha_max} to obtain the final alpha channel value.

\begin{equation}
\label{eq:text_alpha_max}
A_{\text{FG}}' = \max(A_B, A_G, A_R)
\end{equation}

According to the system of equations \ref{eq:text_alpha_equation_group}, it can be observed that when the background color is pure white, i.e., the color channel value is 255, the resulting alpha channel value ranges from negative infinity to zero, making the above equations inapplicable. However, when the background color is pure white, the foreground color is easy to determine, and the color derivation formula can be simplified to equation \ref{eq:alpha_by_white_bg}:

\begin{equation}
\label{eq:alpha_by_white_bg}
A = (255 - ChannelValue) / 255
\end{equation}


\begin{equation}
\label{eq:white_bg_mask}
Mask_{\text{WhiteBG}} = (B_{\text{BG}} = 255) \land (G_{\text{BG}} = 255) \land (R_{\text{BG}} = 255)
\end{equation}

Therefore, we designed a mask extraction approach to find all pixel positions with a pure white background by detecting whether the color channel values are \((255,255,255)\):

\begin{align}
\label{eq:text_alpha_avg}
A_{\text{FG}}' = \frac{1}{6} \Bigg( & \frac{255 - G_{\text{BlueFG}}}{255} + \frac{255 - R_{\text{BlueFG}}}{255} + \frac{255 - B_{\text{GreenFG}}}{255} \nonumber \\
& + \frac{255 - R_{\text{GreenFG}}}{255} + \frac{255 - B_{\text{RedFG}}}{255} + \frac{255 - G_{\text{RedFG}}}{255} \Bigg)
\end{align}

Finally, we apply equation \ref{eq:text_alpha_avg} to the pixels within the mask area and equation \ref{eq:text_alpha_max} to other areas to obtain the alpha channel for all pixels of the target element. With the alpha channel, we can use the following equations \ref{eq:text_bgr_equation_group} to obtain the BGR channel values.

\begin{equation}
\label{eq:text_bgr_equation_group}
\left\{
\begin{aligned}
B_{\text{FG}}' &= \frac{B_{\text{RawFG}} - (1 - A_{\text{FG}}') \cdot B_{\text{BG}}'}{A_{\text{FG}}' + \epsilon} \\
G_{\text{FG}}' &= \frac{G_{\text{RawFG}} - (1 - A_{\text{FG}}') \cdot G_{\text{BG}}'}{A_{\text{FG}}' + \epsilon} \\
R_{\text{FG}}' &= \frac{R_{\text{RawFG}} - (1 - A_{\text{FG}}') \cdot R_{\text{BG}}'}{A_{\text{FG}}' + \epsilon}
\end{aligned}
\right.
\end{equation}

\subsection{Layer Analysis}
\label{subsec:4.3}
In this step, our objective is to obtain the minimum composition layer (\textit{MCL}), which involves assigning all elements in a GUI to different layers such that there is no overlap among elements within each layer, and any element in an upper layer is rendered above elements in a lower layer. Given a set of GUI elements \( E = \{e_1, e_2, \ldots, e_n\} \), each element \( e_i \) is associated with a mask region \( M(e_i) \). We aim to allocate these elements into a set of layers \( L = \{L_1, L_2, \ldots, L_k\} \), where \( k \) is the number of layers, and our goal is to minimize \( k \). Equation \eqref{eq:mcl} describes the constraints of this optimization problem: the union of elements across all layers should equal the entire set of GUI elements \( E \), i.e., \(\bigcup_{i=1}^{k} L_i = E\); for any layer \( L_i \), the mask regions of any two elements \( e_j \) and \( e_l \) should not overlap, i.e., \( M(e_j) \cap M(e_l) = \emptyset \); furthermore, for each pair of adjacent layers \( L_m \) and \( L_{m+1} \), if elements \( e \) and \( e' \) have overlapping mask regions, then the rendering order of \( e' \) must be below that of \( e \). Through these conditions, we define how to efficiently assign GUI elements to the minimum number of layers, thereby obtaining an equivalent Z-axis for the actual composite rendering of the GUI elements.

\begin{equation}
\label{eq:mcl}
\begin{aligned}
\text{Minimize } & \quad k \\
\text{Subject to } & \quad \bigcup_{i=1}^{k} L_i = E, \\
& \quad \forall i, \forall e_j, e_l \in L_i, \, R(e_j) \cap R(e_l) = \emptyset, \\
& \quad \forall m, \forall e \in L_{m+1}, \forall e' \in L_m, \\
& \quad \text{if } R(e) \cap R(e') \neq \emptyset, \, \text{then render order of } e > e'.
\end{aligned}
\end{equation}

We designed an algorithm to obtain the MCL as follows: First, based on the Channel Derivation from the previous step, we have obtained the XY coordinates of each GUI element, allowing us to determine which elements overlap. We then perform pairwise analysis on these overlapping elements. Our approach is inspired by the insight that if we know the stack order of any two elements, we can construct a Directed Acyclic Graph (DAG) and solve for the MCL through topological sorting.

\begin{algorithm}
\caption{Analysis stack order of two GUI elements}
\label{code:compare}
\begin{algorithmic}[1]

\If{\(zindex_i \text{ is an integer} \land zindex_j \text{ is an integer}\)}
    \If{$zindex_i = zindex_j$}
        \If{$w_i \times h_i > w_j \times h_j$}
            \Return $(1, 0)$
        \ElsIf{$w_i \times h_i < w_j \times h_j$}
            \Return $(0, 1)$
        \EndIf
    \ElsIf{$zindex_i < zindex_j$}
        \Return $(1, 0)$
    \Else
        \Return $(0, 1)$
    \EndIf
\Else
    \If{$w_i \times h_i > w_j \times h_j$}
        \Return $(1, 0)$
    \ElsIf{$w_i \times h_i < w_j \times h_j$}
        \Return $(0, 1)$
    \EndIf
\EndIf

\end{algorithmic}
\end{algorithm}

We use binary vectors to represent the stacking order of any two GUI elements: 1. If element \( e_i \) is above element \( e_j \), we use the vector \( \mathbf{v}_{ij} = (1, 0) \) to represent this order. 2. Conversely, if element \( e_j \) is above element \( e_i \), the vector is \( \mathbf{v}_{ji} = (0, 1) \).

According to the rendering mechanism of HTML, descendant nodes are above the rendering level of parent nodes. Therefore, we only need to compare two overlapping GUI elements that are not on the same XPath. Code \ref{code:compare} shows the comparison of two GUI elements through Z-Index and area. First, we compare their Z-Index attributes. If both GUI elements have numeric Z-Index values, we determine the stacking order based on the size of the Z-Index. If the Z-Index attribute is meaningless, such as auto, or if the Z-Index values of the two GUI elements are the same, we determine their hierarchy by comparing the areas of the two GUI elements. Based on our experience, larger elements are usually below smaller elements.

For all elements on the page \( E = \{e_1, e_2, \ldots, e_n\} \), we can construct all possible binary vectors to represent their stacking order. Additionally, for elements in the set \( E \), if certain elements do not overlap with any other elements, these elements can be considered as isolated vertices. These isolated vertices have no edges connecting them to other nodes in the DAG, indicating that their stacking order is independent of other elements:

\begin{equation}
\label{eq:dag}
V = \left\{ \mathbf{v}_{ij} \mid e_i, e_j \in E, \, i \neq j \right\} \cup \{e_k \mid e_k \in E, \, e_k \text{is isolated vertex}\}
\end{equation}

Finally, we apply Kahn's Algorithm for topological sorting on the DAG represented by Equation \ref{eq:dag} to obtain the MCL that minimizes \( k \). Each layer of the MCL records all elements present in that layer.5

\subsection{Output Annotations}
\label{subsec:4.4}

We process each HTML node on the page through the channel derivation and layer analysis modules, obtaining the BGRA bitmap of each GUI element, as well as the accurate spatial coordinates (X, Y, Z) and size information (W, H) of the GUI elements. Based on this information, we derive the annotations for the Mask, Contour, and BBox.



\begin{enumerate}
    \item \textbf{Mask}: First, define the BGRA bitmap of the target element as matrix \( M \), where the \(\alpha\) channel value of each pixel is \( M(x, y, \alpha) \). We set all pixels satisfying \( M(x, y, \alpha) \neq 0 \) to color \( C_1 \), while pixels of other elements are set to color \( C_2 \), where \( C_2 \) applies to all non-target elements with non-zero \(\alpha\) channel values.Next, recompose according to the layer order using Equation \ref{eq:color_render} to obtain a GUI page image, denoted as \(\text{Element}_i^{\text{Visible}}\).Then, compose another GUI page image excluding the visible part of the target element, denoted as \(\text{Element}_i^{\text{Invisible}}\).Finally, calculate the difference between the two images to get mask:
    
    \[
    D = \left| \text{Element}_i^{\text{Visible}} - \text{Element}_i^{\text{Invisible}} \right|
    \]

\( D \) is the mask of the element's visible area.
    \item \textbf{Contour}: We find the outer contour of all disconnected regions in the mask to obtain the contour.5
    \item \textbf{BBox}: We calculate the minimum enclosing rectangle of the mask to obtain the BBox.
\end{enumerate}

\section{Implementation}
\begin{figure*}[htbp]
    \centering
    \begin{subfigure}[b]{0.19\textwidth}
        \centering
        \includegraphics[height=0.25\textheight]{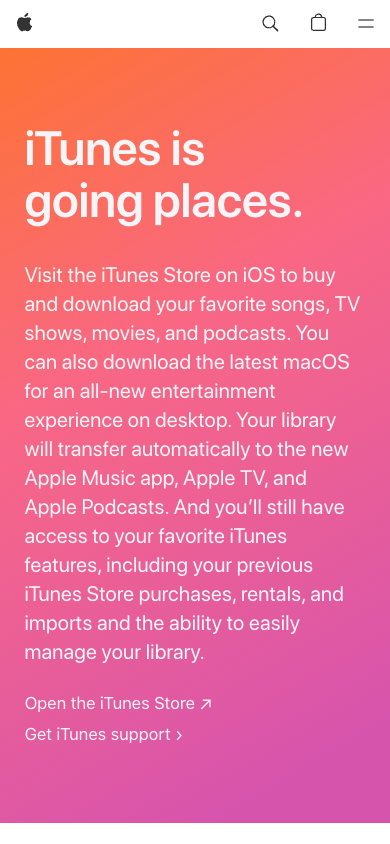}
        \caption{GUI screenshot}
    \end{subfigure}
    \begin{subfigure}[b]{0.19\textwidth}
        \centering
        \includegraphics[height=0.25\textheight]{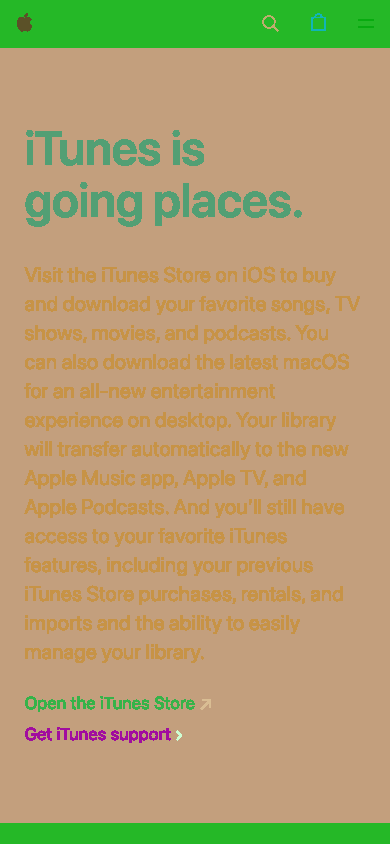}
        \caption{PixelWeb mask}
    \end{subfigure}
    \begin{subfigure}[b]{0.19\textwidth}
        \centering
        \includegraphics[height=0.25\textheight]{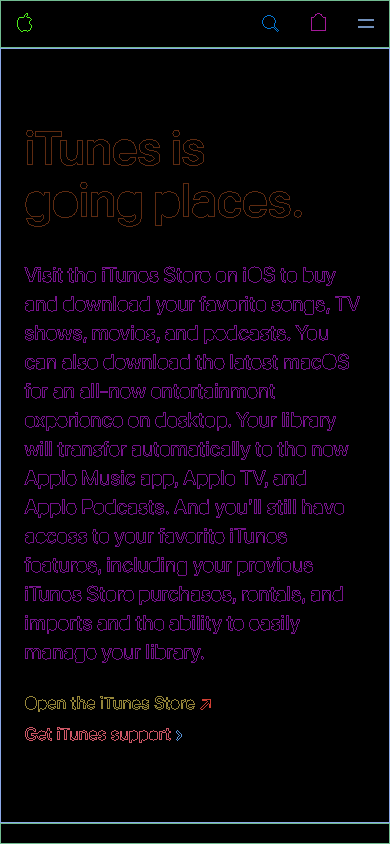}
        \caption{PixelWeb contour}
    \end{subfigure}
    \begin{subfigure}[b]{0.19\textwidth}
        \centering
        \includegraphics[height=0.25\textheight]{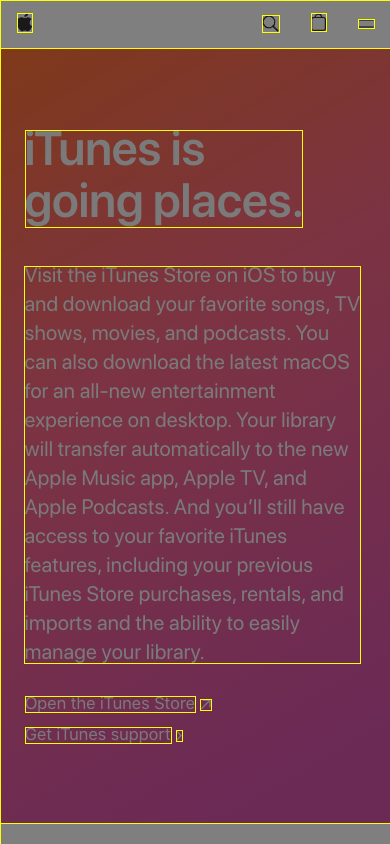}
        \caption{PixelWeb BBox}
    \end{subfigure}
    \begin{subfigure}[b]{0.19\textwidth}
        \centering
        \includegraphics[height=0.25\textheight]{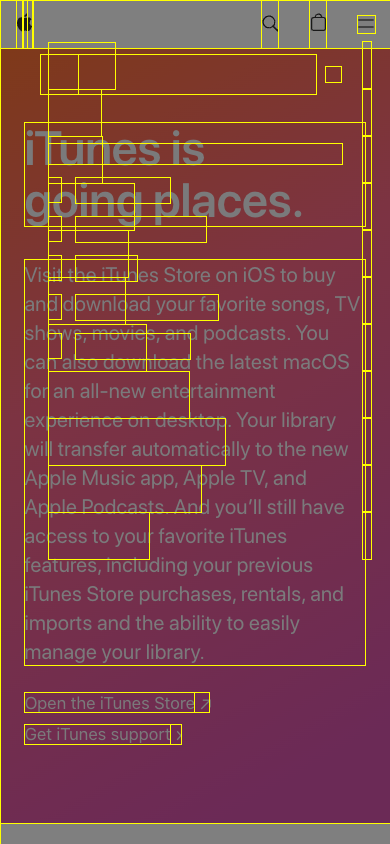}
        \caption{WebUI BBox}
    \end{subfigure}
    \caption{Example 1 of annotations in PixelWeb and WebUI}
    \label{fig:dataset_examples_1}
\end{figure*}

\begin{figure*}[htbp]
    \centering
    \begin{subfigure}[b]{0.19\textwidth}
        \centering
        \includegraphics[height=0.25\textheight]{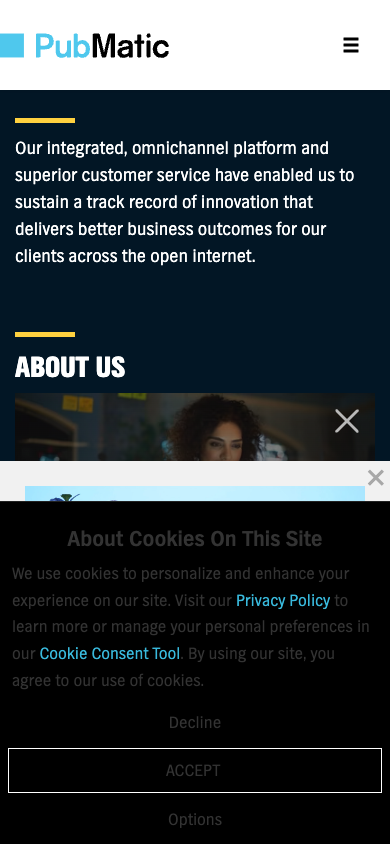}
        \caption{GUI screenshot}
    \end{subfigure}
    \begin{subfigure}[b]{0.19\textwidth}
        \centering
        \includegraphics[height=0.25\textheight]{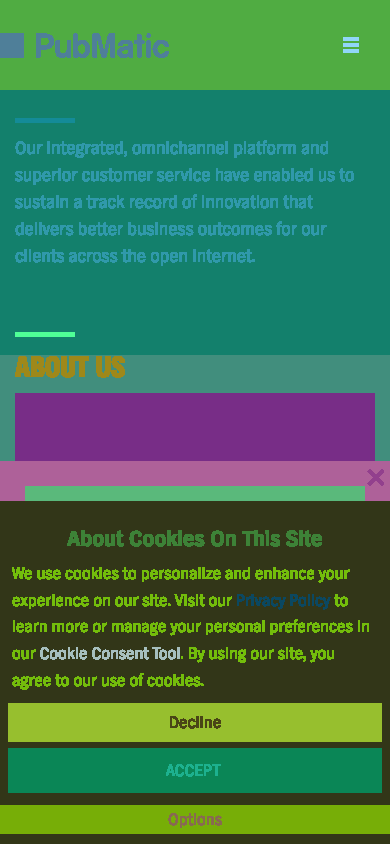}
        \caption{PixelWeb Mask}
    \end{subfigure}
    \begin{subfigure}[b]{0.19\textwidth}
        \centering
        \includegraphics[height=0.25\textheight]{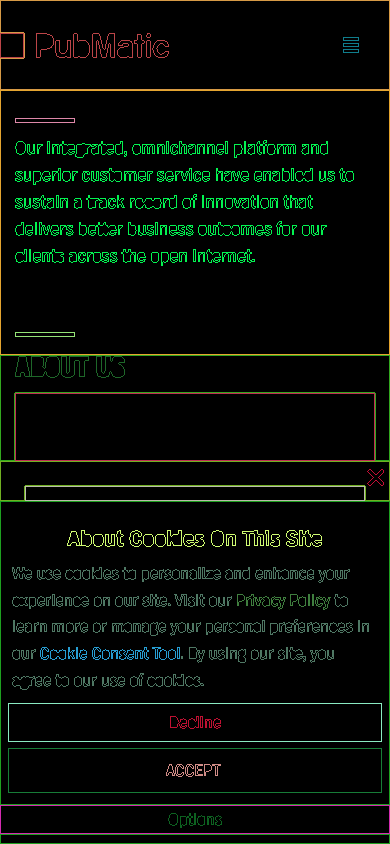}
        \caption{PixelWeb Contour}
    \end{subfigure}
    \begin{subfigure}[b]{0.19\textwidth}
        \centering
        \includegraphics[height=0.25\textheight]{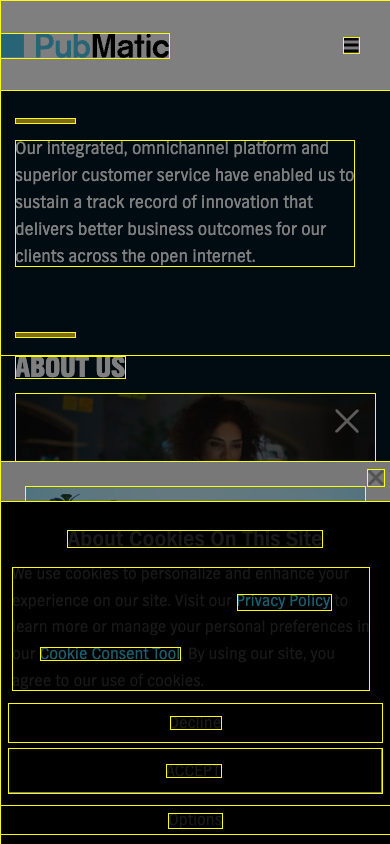}
        \caption{PixelWeb BBox}
    \end{subfigure}
    \begin{subfigure}[b]{0.19\textwidth}
        \centering
        \includegraphics[height=0.25\textheight]{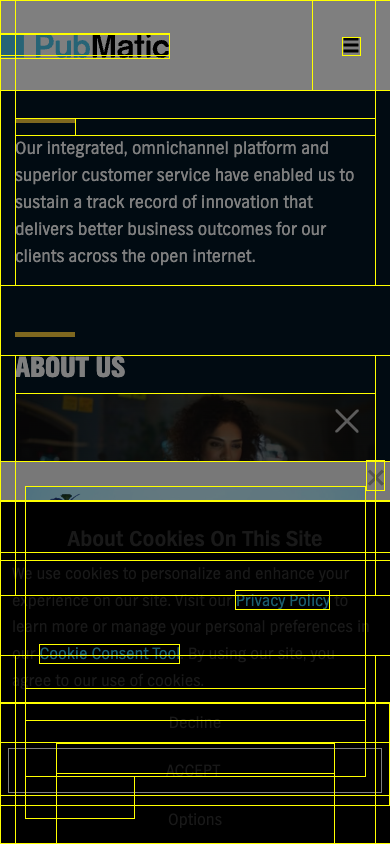}
        \caption{WebUI BBox}
    \end{subfigure}
    \caption{Example 2 of annotations in PixelWeb and WebUI}
    \label{fig:dataset_examples_2}
\end{figure*}

\begin{figure*}[htbp]
    \centering
    \begin{subfigure}[b]{0.19\textwidth}
        \centering
        \includegraphics[height=0.25\textheight]{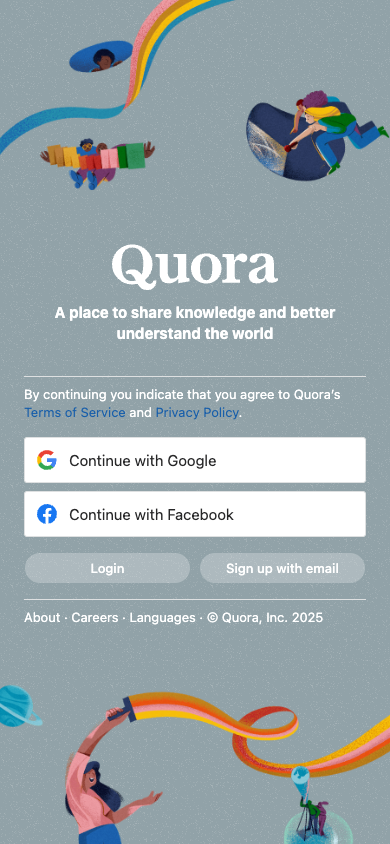}
        \caption{GUI screenshot}
    \end{subfigure}
    \begin{subfigure}[b]{0.19\textwidth}
        \centering
        \includegraphics[height=0.25\textheight]{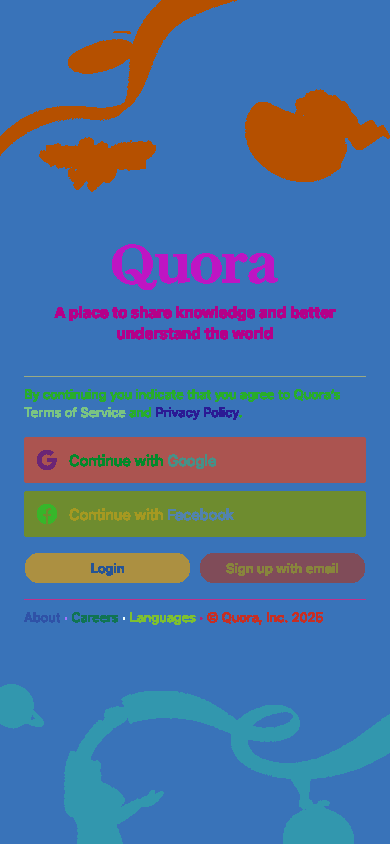}
        \caption{PixelWeb Mask}
    \end{subfigure}
    \begin{subfigure}[b]{0.19\textwidth}
        \centering
        \includegraphics[height=0.25\textheight]{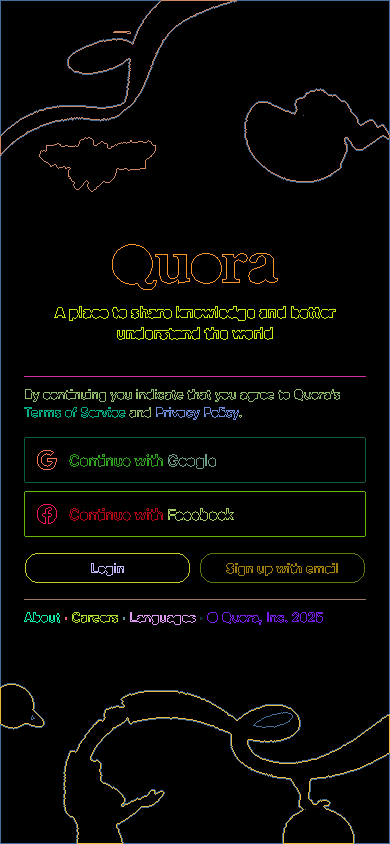}
        \caption{PixelWeb Contour}
    \end{subfigure}
    \begin{subfigure}[b]{0.19\textwidth}
        \centering
        \includegraphics[height=0.25\textheight]{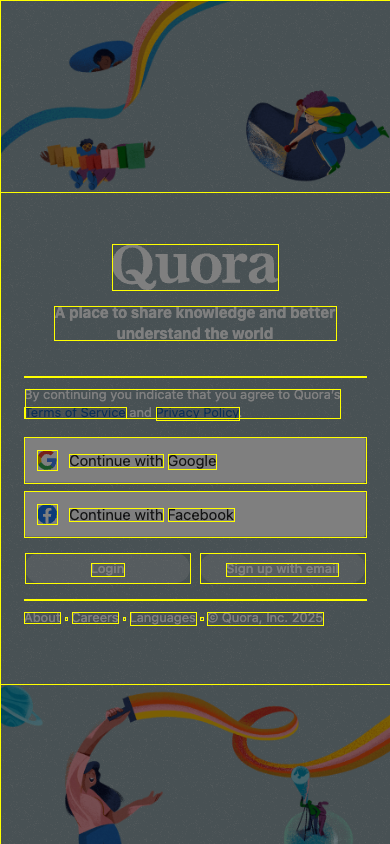}
        \caption{PixelWeb BBox}
    \end{subfigure}
    \begin{subfigure}[b]{0.19\textwidth}
        \centering
        \includegraphics[height=0.25\textheight]{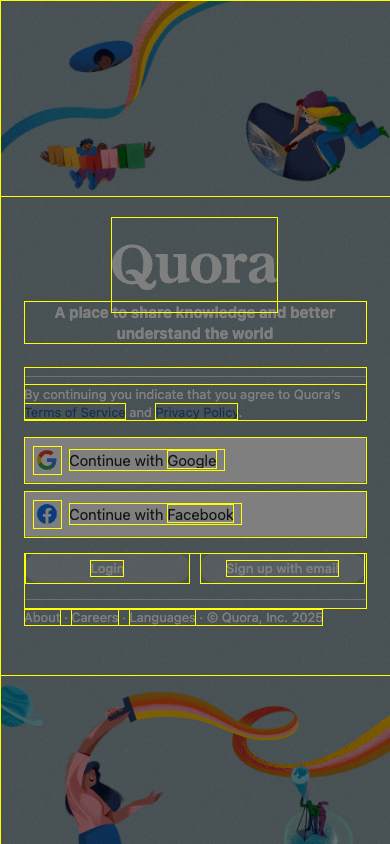}
        \caption{WebUI BBox}
    \end{subfigure}
    \caption{Example 3 of annotations in PixelWeb and WebUI}
    \label{fig:dataset_examples_3}
\end{figure*}

The implementation of our approach primarily includes three components: page crawler, page operation, and annotation export.

The page crawler component is implemented based on package playwright~\cite{playwright}. We crawled each webpage corresponding to URLs from the WebUI~\cite{Wu2023WebUI} dataset. We skipped URLs without actual contents. During crawling, we filtered out pages with a 404 status by simply determining if the number of page nodes was less than 5 and all page elements were concentrated at the top of the screen. Since our approach requires repeated operations on page nodes, pages need to be re-rendered repeatedly, and the overhead for capturing screenshots is also high. So we performed preliminary filtering of nodes. We filtered elements based on CSS properties to reduce the workload of subsequent steps. Our filtering criteria are as follows. 1) If CSS properties such as display, visibility, and opacity are set to be invisible, we consider the element invisible. 2) If the element's getBoundingClientRect is outside the visible area, we consider it invisible. Based on our observations, the bounding box obtained by getBoundingClientRect is generally larger rather than smaller, making it reliable for preliminary judgment of element visibility. 3) If elements have background or innerText, we preliminarily consider them visible. After this filtering, the number of nodes to be processed is significantly reduced, thereby improving the efficiency of the entire implementation.

\begin{figure*}[htbp]
  \centering
  \begin{subfigure}[b]{0.33\textwidth}
    \includegraphics[width=\textwidth]{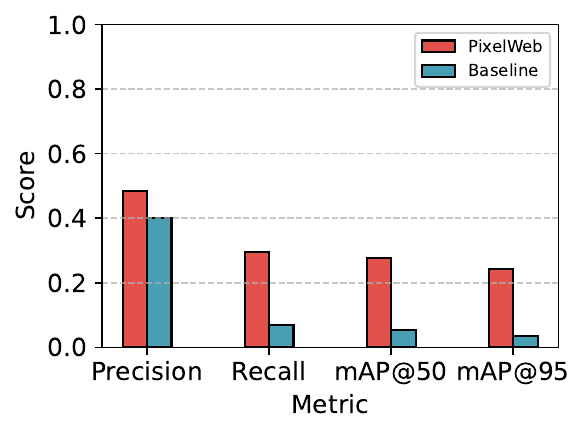}
    \caption{2.6M parameters with class}
  \end{subfigure}
  \hfill
  \begin{subfigure}[b]{0.33\textwidth}
    \includegraphics[width=\textwidth]{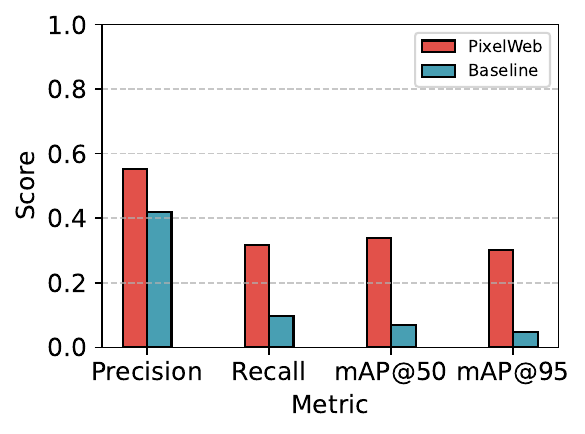}
    \caption{9.3M parameters with class}
  \end{subfigure}
  \hfill
  \begin{subfigure}[b]{0.33\textwidth}
    \includegraphics[width=\textwidth]{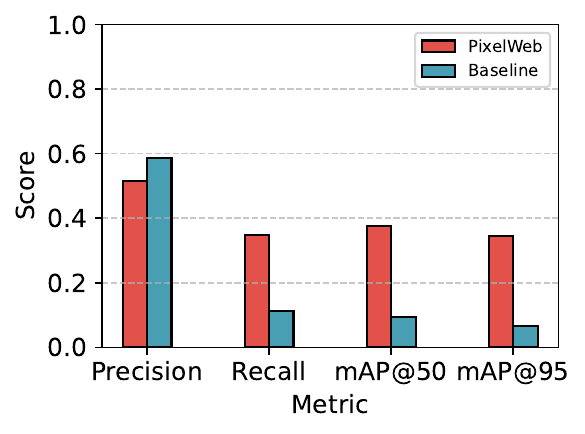}
    \caption{20.2M parameters with class}
  \end{subfigure}

  \vspace{0.5em}

  \begin{subfigure}[b]{0.33\textwidth}
    \includegraphics[width=\textwidth]{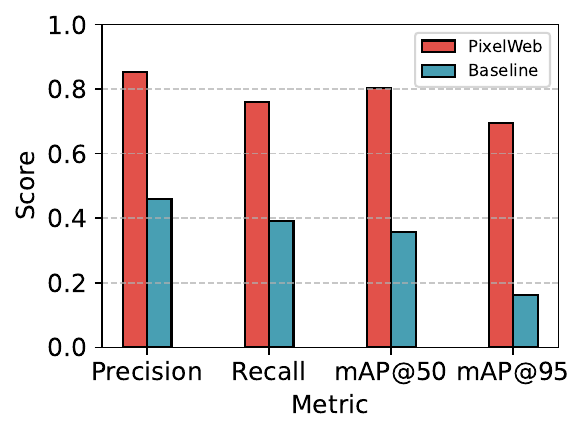}
    \caption{2.6M parameters w/o class}
  \end{subfigure}
  \hfill
  \begin{subfigure}[b]{0.33\textwidth}
    \includegraphics[width=\textwidth]{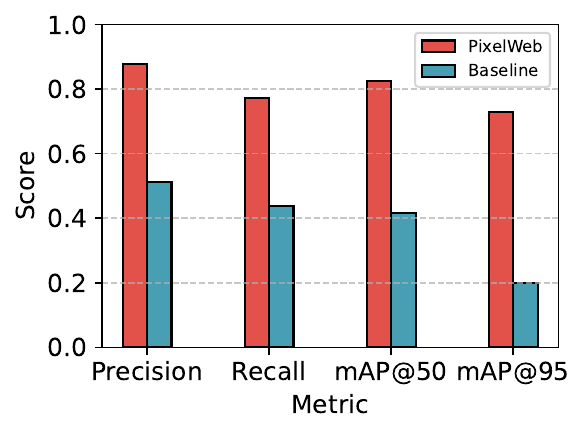}
    \caption{9.3M parameters w/o class}
  \end{subfigure}
  \hfill
  \begin{subfigure}[b]{0.33\textwidth}
    \includegraphics[width=\textwidth]{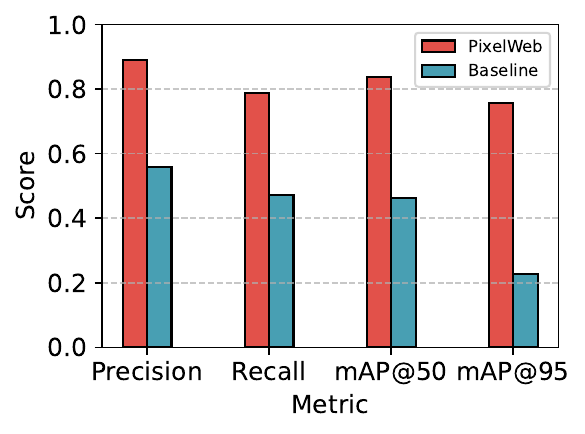}
    \caption{20.2M parameters w/o class}
  \end{subfigure}

  \caption{Evaluation results for different model sizes with and without class labels}
  \label{fig:main_experiment}
\end{figure*}

The page operation component manipulates the page using JavaScript to achieve DOM transformations and metadata retrieval required at each step of the approach. This component is implemented using the evaluate function provided by playwright to inject JavaScript code. According to HTML5 specifications and browser rendering principles, we inject desired styles by modifying the inline style of elements and adding the !important attribute. We maintain a mapping table to record the original inline style of elements, allowing us to restore the original style when repeated transformations are needed. For hiding and showing elements, we use the visibility property. For pseudo-elements, we control their visibility using opacity and treat them as graphics. To change the GUI background color, we set the background color on the HTML root tag. For text color changes, we set the color property.

The annotation export component is implemented based on Pillow~\cite{pillow} and OpenCV~\cite{opencv}. We use the paste function of Pillow for layer composition and obtain the mask through the absdiff provided by OpenCV combined with matrix operations. Then, we obtain the contour using the findContours function.

Finally, we construct the PixelWeb dataset containing 100,000 samples. Each data sample includes a screenshot of a webpage GUI, along with transparent background images of each GUI element, XYZ axis positions, masks, contours, BBox, and corresponding code and computed style information. Figure~\ref{fig:dataset_examples_1}-\ref{fig:dataset_examples_3} show examples of our data samples and examples of annotating the same samples using the WebUI approach.

\section{Evaluation}
In this section, we first assess the quality of the PixelWeb generated by our proposed method through a user study (Section~\ref{sec:annotation_quality}). We then demonstrate the effectiveness of PixelWeb on various GUI downstream tasks in Section~\ref{sec:Results_on_Downstream_Tasks}.

\subsection{Setup}

\subsubsection{Models}
We use the YOLOv12~\cite{tian2025yolov12,yolo12} series models released in 2025 for experiments on downstream tasks. We try three versions with different parameter scales: 2.6M, 9.3M, and 20.2M, to explore how the dataset performs on different model scales. When training different models and datasets, we set the batch size uniformly to 16, and all other parameters use the default configuration of YOLOv12. In terms of metrics, we use precision, recall, and mAP metrics, distinguishing between mAP@50 and mAP@95.

\subsubsection{Baseline}
To ensure the fairness of the experiment, while constructing the PixelWeb dataset, we use the annotation method of WebUI to annotate the same data samples, including both BBox and classification annotations, as the baseline for the experiment. The only difference between the PixelWeb and baseline datasets is the method of BBox annotation, allowing for a fair comparison to determine if our annotation method effectively improves BBox quality. We construct PixelWeb and baseline datasets containing a total of 10,000 samples, with 9,000 samples in the training set and 1,000 samples in the validation set. To get an totally precise annotation, we additionally annotate 200 samples to create a test set.

\subsection{Annotation Quality}
\label{sec:annotation_quality}
\begin{table}[]
\caption{Results of User Study}
\label{tab:user_study}
\begin{tabular}{l|ccccc}
\hline
 & User1 & User2 & User3 & Total \\ \hline
PixelWeb is much better & 85 &92 & 61 & 238 \\
PixelWeb is better & 102 & 58 & 99 & 259 \\
Not sure & 53 & 100 & 88 & 241 \\
Baseline is better & 39 & 24 & 39 & 102 \\
Baseline is much better & 21 & 26 & 13 & 60 \\

\hline

\end{tabular}
\end{table}

We created 300 slides, each featuring two GUI images with overlaid BBoxes, arranged side by side. The images are randomly selected from the PixelWeb and Baseline datasets, with their order randomized. We engaged three participants with software engineering backgrounds, explained the types of BBox annotation errors, and clarified the evaluation criteria. The participants were then asked to annotate the slides, with options for each: the left sample is slightly better, the left sample is much better, about the same, the right sample is slightly better, or the right sample is much better.

Table \ref{tab:user_study} shows the results of the user study. PixelWeb received a significantly higher number of "much better" ratings, accounting for 26\% of all samples. Overall, 53\% of the samples were rated better than the baseline, while only 18\% were rated worse. Thus, PixelWeb is subjectively considered superior to the existing baseline.
\subsection{Results on Downstream Tasks}
\label{sec:Results_on_Downstream_Tasks}

Figure \ref{fig:main_experiment} display the training results for parameter sizes of 2.6M, 9.3M, and 20.2M, respectively. We observe that across various parameter scales and classification labels, PixelWeb consistently outperforms the Baseline in metrics. As model size increases, advantage of PixelWeb remains significant and stabilizes.

Although under the 20.2M model parameters, the Precision metric of PixelWeb is slightly weaker than that of the Baseline when trained with classification labels, the Recall and mAP metrics are significantly better than those of the Baseline. Therefore, overall, PixelWeb is still significantly better than the Baseline. Moreover, as the scale of model parameters increases exponentially, the improvement of the Baseline is not obvious. Even when training the Baseline with the model having the largest parameters in the experiment, all metrics are significantly lower than those of PixelWeb trained with only 1/10 of the parameters. Since the only difference between PixelWeb and the Baseline is the accuracy of BBox annotations, this indicates that the higher quality BBox annotations of PixelWeb indeed lead to a significant improvement in the training effect of the object detection task. Given that the primary difference between PixelWeb and Baseline is the accuracy of BBox annotations, this suggests that the higher quality of annotations in PixelWeb significantly enhances training performance for object detection tasks.
\section{Potential for GUI Tasks}
\begin{table}[t]
\caption{Expanding GUI downstream tasks with PixelWeb new annotations}
\label{tab:pweb_power}
\centering
\begin{tabularx}{0.47\textwidth}{|l|X|}
\hline
\textbf{PixelWeb New Annotation} & \textbf{New Tasks} \\ \hline
\multirow{4}{*}{\centering Element Image} & GUI element retrieval \\ \cline{2-2}
                                          & GUI element generation \\ \cline{2-2}
                                          & GUI layout generation by images \\ \cline{2-2}
                                          & GUI composting dataset \\ \hline
\multirow{2}{*}{\centering Element Layer} & GUI layout with Z-axis direction \\ \cline{2-2}
                                          & Advertisement element analysis \\ \hline
\multirow{4}{*}{\centering Element Mask} & GUI instance segmentation \\ \cline{2-2}
                                         & GUI semantic segmentation \\ \cline{2-2}
                                         & GUI element shape analysis \\ \cline{2-2}
                                         & GUI element color generation \\ \hline
\multirow{3}{*}{\centering Element Contour} & GUI instance segmentation \\ \cline{2-2}
                                            & GUI semantic segmentation \\ \cline{2-2}
                                            & GUI element shape analysis \\ \hline
\multirow{3}{*}{\centering Element Computed Style} & GUI element generation \\ \cline{2-2}
                                                   & GUI element image2code \\ \cline{2-2}
                                                   & GUI element code2image \\ \hline
\end{tabularx}
\vspace{-10pt} 
\end{table}

PixelWeb shows potential in a wide range of GUI downstream tasks, as shown in Table \ref{tab:datasets_diff}. Accurate BBox annotation is just one of the many labels we provide. Compared to existing datasets, we offer richer annotation information at the element level, including element image and element mask. These new types of annotation information not only enhance various existing GUI downstream tasks but also enable various new GUI downstream tasks. Table \ref{tab:pweb_power} demonstrates the new GUI downstream tasks that these novel labels enable. In detection and generation tasks, due to our provision of higher-dimensional information annotations, more fine-grained tasks can be achieved, such as element-level generation tasks and semantic segmentation tasks.

\section{Conclusion}
In this paper, we proposed a novel automated annotation method resulting in the creation of the PixelWeb dataset. This dataset includes BGRA four-channel bitmap annotations and layer position annotations for GUI elements, along with precise bounding box, contour, and mask annotations. Our experimental results demonstrate that PixelWeb significantly outperforms existing GUI datasets, leading to enhanced performance in downstream tasks such as GUI element detection. Looking forward, this dataset can enhance performance in a wider range of tasks, such as intelligent agent interactions and automated web page generation. Furthermore, exploring the potential of transferring these techniques to mobile apps and other non-web GUIs could provide even greater versatility and impact.

\bibliography{reference}
\bibliographystyle{plain}

\end{document}